\documentclass[9pt,twocolumn,twoside]{pnas-new}

\templatetype{pnasresearcharticle} 

\usepackage{multirow}

\title{Muscle-inspired flexible mechanical logic architecture for colloidal robotics}

\author[a]{Mayank Agrawal}
\author[a,b,c, $\ast$]{Sharon C Glotzer} 

\affil[a]{Department of Chemical Engineering, University of Michigan, Ann Arbor, Michigan 48109-2136, USA}
\affil[b]{Department of Materials Science and Engineering, University of Michigan, Ann Arbor, Michigan 48109-2136, USA}
\affil[c]{Biointerfaces Institute, University of Michigan, Ann Arbor, Michigan 48109-2136, USA}

\leadauthor{Agrawal} 

\authordeclaration{Authors declare no conflict of interest.}
\correspondingauthor{To whom correspondence should be addressed. E-mail: sglotzer@umich.edu}


\begin{abstract}
Materials that respond to external stimuli by expanding or contracting provide a transduction route that integrates sensing and actuation powered directly by the stimuli. This motivates us to build colloidal scale robots using these materials that can morph into arbitrary configurations. For intelligent use of global stimuli in robotic systems, computation ability needs to be incorporated within them. The challenge is to design an architecture that is compact, material agnostic, stable under stochastic forces and can employ stimuli-responsive materials. We present an architecture that computes combinatorial logic using mechanical gates that use muscle-like response -- expansion and contraction -- as circuit signal with additional benefits of logic circuitry being physically flexible and able to be retrofit to arbitrary robot bodies. We mathematically analyze gate geometry and discuss tuning it for the given requirements of signal dimension and magnitude. We validate the function and stability of the design at the colloidal scale using Brownian dynamics simulations. We also demonstrate the gate design using a 3D printed model. Finally, we simulate a complete robot consisting of its mechanical circuit connected to the robot’s skeleton that folds into Tetris shapes.
\end{abstract}

\begin{document}

\maketitle
\thispagestyle{firststyle}
\ifthenelse{\boolean{shortarticle}}{\ifthenelse{\boolean{singlecolumn}}{\abscontentformatted}{\abscontent}}{}

\section{Introduction}

Active materials such as stimuli-responsive polymers (SRPs)\cite{Wei2017,Hu2012,Ghosh2017} and chains of patchy particles\cite{Shah2015} harness environmental stimuli to contract or expand, thereby behaving as \emph{artificial muscles}. There is an interest in the community to use such materials to build robots that can morph their structure in response to external stimuli. The vision is to create robotic systems that can react to their environment organically and do not particularly rely on electronic components. Examples of such robotic systems include untethered grippers\cite{Ghosh2017}, reconfigurable structures\cite{Mao2016}, origami/kirigami sheets\cite{Tang2019}, and smart textiles\cite{Hu2012}. A large variety of SRPs that are available to harness stimuli such as temperature, pH, humidity, light, sound, chemical gradient, mechanical force, and electric/magnetic fields provide a rich design space of materials and energy sources to build such robots\cite{Wei2017,Hu2012,Ghosh2017, Dong2019}. A particular opportunity arises from employing multiple stimuli in a single robotic system to advance its intelligence. One such advancement is the ability to compute. The idea is that different stimuli will incur different SRP responses that will be processed by the computing architecture built into the robotic system. The processed information can then be used by the robot as required. Current computing architectures that use SRPs perform the computation for advanced analysis; the stimulus input is processed by a mechanical logic circuit and a chemical response is generated as the output\cite{Zhang2019}. In contrast, our goal is to process the stimulus input to generate a mechanical output that can then be used to morph or move the robot. The motivation behind this goal is that to accomplish a given task, it may be required to modify the robot's structure into multiple configurations. A logic architecture provides a systematic and modular way to achieve these configurations using only a handful of stimuli. In principle, $2^N$ configurations can be achieved using $N$ different stimuli via binary logic computation.

Our other goal is to make this logic architecture compatible with colloidal scale physics so as to build intelligent robots at this scale. To realize both goals, here we present a mechanical logic architecture consisting of gate mechanisms and signal transmission mechanisms to perform logic computation. Our design innovation in these mechanisms is to use expansion and contraction as the mechanical signal rather than linear displacement, which is usually used, in order to increase compatibility with SRPs, which also respond by contracting or expanding. Additionally, our gate mechanisms are 3D in contrast to the usual 2D ones\cite{Zhang2017,Song2019}. These two features enabled the construction of the logic gates, which are devoid of any colliding/sliding parts and are relatively more compact as compared to previous mechanical logic architectures\cite{Zhang2017,Song2019}. Our other innovation is the flexible connector design to transmit the signal to and from the gates. The flexible connector, in turn, forms flexible logic circuits. Thus, the novelty of our architecture is three-fold: 1) Usage of expansion/contraction, rather than translation, as the signal. 2) Compact gates without colliding/sliding parts that are stable under colloidal-scale stochastic forces. 3) Flexible circuitry. 

We first introduce the design of the mechanical \emph{logic core}, which forms the base structure of all the logic gates. We then discuss using this logic core along with levers to construct the NAND gate (universal logic gate) and subsequently other logic gates. Next, we describe the connector design and subsequently analyze the frequency response and information loss for a simple circuit. Finally, we present a proof-of-concept demonstration of a complete robot consisting of a mechanical logic circuit connected to a skeleton chain that folds into four different Tetris shapes using inputs generated via two different "muscles." We also provide a table (Table 1) of existing technologies for fabricating and actuating complex miniature structures, and their associated material properties.

\begin{figure}[t]
\centering
\includegraphics[width=1\linewidth]{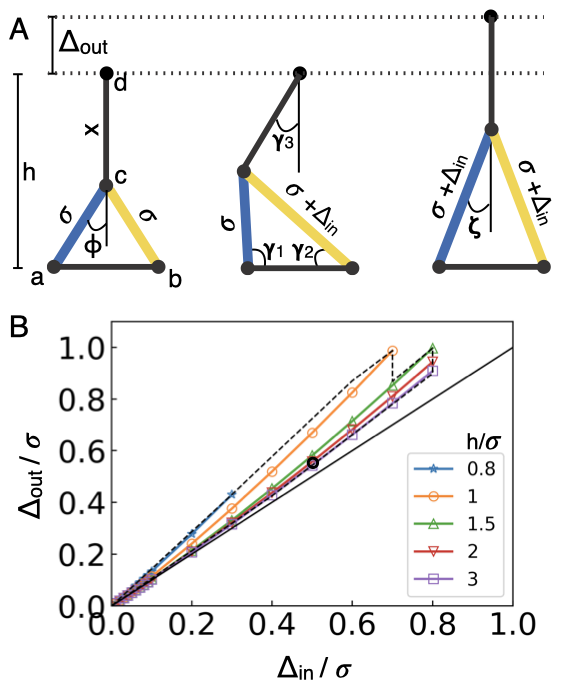}
\caption{Mechanical logic core. (A) Configurations of the 2D link mechanism referred to as of the \emph{logic core} for different actuation states of the blue and yellow links (muscle links). The constraint on the mechanism is such that the point $d$ moves along the perpendicular bisector of the link $ab$. Given $\sigma$, $\Delta_{in}$, and the height of the logic core defined as $h=x+\sigma \cos{\phi}$, the mechanism geometry can be calculated such that both the muscle links must expand by $\Delta_{in}$ for the point to $d$ to raise by $\Delta_{out}$, and perform the logic computation. Solid circles are the hinges. Black lines are fixed length links. (B) Plot of $\Delta_{out}$ vs $\Delta_{in}$ for different $h$. The dotted line is the boundary of accessible values of $\Delta_{out}$ and $\Delta_{in}$. The black hollow circle points to the parameter values used in this work, which are $\Delta_{in}/\sigma = 0.5$ and $h/\sigma=2$ yielding $\Delta_{out}/\sigma = 0.56$. Black solid line has slope one. } 
 
\label{figure1}
\end{figure}

\section{Mechanical logic core}
The logic operation of the 3D gates is implemented by a 2D mechanism, which we refer to as the \emph{logic core}, shown in Figure 1A. The mechanism consists of two passive links -- $ab$ and $cd$ -- and two active, or \emph{muscle} links -- $ac$ and $bc$ -- shown in blue and yellow colors respectively. The links are connected at joints and are free to rotate in the plane. The only constraint in this mechanism is that point $d$ moves along the perpendicular bisector of $ab$. The muscle links expand to increase lengths $ac$ and $bc$ from $\sigma$ to $\sigma+\Delta_{in}$. The geometry of the mechanism can be optimized such that point $d$ rises, by a distance $\Delta_{out}$, only when both the lengths $ac$ and $bc$ are elongated by $\Delta_{in}$ (see Figure 1A). This geometrical property results in the logic operation of the mechanism. The geometry of the mechanism is defined by the parameters $x/\sigma$ and $\phi$. These parameters can be calculated using the set of equations in Eq. 1, 
\begin{equation}
\begin{aligned}
\sin \phi - \cos \gamma_1 - (x/\sigma) \sin \gamma_3 &=0\\
\sin \gamma_1 + (x/\sigma) \cos \gamma_3 - x/\sigma - \cos \phi &=0\\
\cos \gamma_1 + (1+\Delta_{in}/\sigma) \cos \gamma_2 - 2\sin \phi &=0\\
\sin \gamma_1 - (1+\Delta_{in}/\sigma) \sin \gamma_2 &=0\\
(1+\Delta_{in}/\sigma) \cos{\zeta} - \cos \phi -\Delta_{out} &=0\\
\sin{\phi}/(1+\Delta_{in}/\sigma) - \sin{\zeta} &=0,
\end{aligned}
\end{equation}
for the given value of quantities $\Delta_{in}/\sigma$, fractional increase in the muscle lengths, and $h/\sigma$, height of the mechanism relative to $\sigma$, where $h=x+\sigma \cos{\phi}$. $\gamma$s and $\zeta$ are internal angle variables. Figure 1B plots $\Delta_{out}$ with respect to $\Delta_{in}$ for various $h$ showing the space of achievable $\Delta_{out}$. The geometry of the logic core is fixed for all the gates.

\section{Gate design}

The logic gates are constructed using the logic core and scissor lever mechanism. The input and output of the logic gates are defined by their rest lengths and signal. The rest length is the distance between the points along which the expansion/contraction (signal) occurs to go from state 0 to state 1. For the sake of consistency, the rest lengths are fixed to $\sigma$ and the signal to $\sigma/2$ for both expansion and contraction. To construct a logic gate, we extend the logic core mechanism into its plane to form slabs from links and attach three scissor levers to its inputs and output as shown in Figure 2A for the NAND gate. The two muscles at $ac$ and $bc$, shown in Figure 1A, are then replaced by two levers. In turn, the muscles are now attached to the inputs of these levers. For this example, we are using contracting muscles. The third lever is connected along the third dimension, which is into the plane of the logic core. The input length of this lever is the distance of $d$ from $ab$. The output of this lever is the output of the gate (see Movie S1 for the simulation). For visualization purposes, we project the 3D gate structure into 2D while retaining the relative scaling of the geometric features (shown in Figure 2A). The purpose of the levers is three-fold:
\smallskip

1) \emph{Impose geometric constraint} - The levers exhibit planar motion. Attaching them to the slabs corresponding to $ab$ and $cd$ using hinges (see Figure 2A) ensures that these slabs rotate in the plane of the logic core. The third lever is attached to the logic core via the hinge at the bottom and the universal joint at the top. This ensures that the motion of the point $d$ perpendicular to $ab$ satisfies the constraint of the logic core mechanism.
\smallskip

2) \emph{Invert signal from expansion to contraction and vice-versa as needed} - The two types of lever mechanisms used here are shown in Figure 2B. By design, the lever on the left inverts contraction into expansion and vice-versa, whereas the one on the right retains the nature of the signal. These two mechanisms are represented by {\it open} and {\it crossed} hinges in Figure 2B. The NAND gate in Figure 2A uses levers with the open hinge mechanism to invert the contraction of the muscle into expansion, which is then fed as input to the logic core. The output of the NAND gate is in the contracted state (state 1) when the inputs are at rest (state 0). When the gate inputs shift to state 1, point $d$ rises, elongating its distance from slab $ab$. This elongation is input to the third lever as the expansion signal. To use this expansion to shift the output of this lever from contracted to rest, the crossed hinge lever mechanism is used. Hence, in this case, the third lever is also acting as a NOT gate.
\smallskip

3) \emph{Act as an adaptor between input and output of different rest lengths and signals} - For example, the input of the third lever has a rest length of $h$ and signal of $0.5\sigma$ expansion. The lever adapts this to the output with a rest length of $\sigma$ and signal of $0.5\sigma$. This adaptation is important when connecting different gates to form a circuit. The geometry is defined by the angles $\theta_1$ and $\theta_2$, and the lengths $l_1$ and $l_2$ as shown in Figure 2B. These parameters can be calculated using the set of coupled equations in Eq. 2 given the values of ($l_{in}$, $l_{out}$), the input and output rest lengths; their expansions ($\Delta l_{in}$, $\Delta l_{out}$), the input and output signals; and $l$, the horizontal length of the lever when its input length is rest length ($l_{in}$). We consistently use $l=1.5 \sigma$ for the input levers -- levers attached to the logic core's inputs -- and $l=2\,\sigma$ for the output lever -- lever attached to the logic core's output.

\begin{equation}
\begin{aligned}
l_1 \sin \theta_1 - l_{in} &= 0\\
l_2 \sin \theta_2 - l_{out} &= 0\\
l_1 \sin (\theta_1+\Delta \theta) - (l_{in}+\Delta l_{in})/2 &= 0\\
l_2 \sin (\theta_2 \pm \Delta \theta) - (l_{out}+\Delta l_{out})/2 &= 0\\
l_1 \cos \theta_1 + l_2 \cos \theta_2 - l&= 0,
\end{aligned}
\end{equation}
where $+$ is used for the crossed hinge lever and $-$ for the open hinge lever. Figure 2C shows the truth table of the NAND gate in Figure 2A containing snapshots of the simulation (from Movie S1). For the simulation, the slabs are rigid and free to rotate about their hinges. The muscles contract or expand when actuated but are stiff otherwise. For the NAND gate in Figure 2A,C, the blue and yellow muscles contract when actuated, defining the binary states 0 (rest) and 1 (contracted). The separate colors for muscles indicate that muscles can be independently actuated, referring to using different stimuli to actuate different SRPs. The motion of our gate mechanism is continuous and, in principle, there is no potential energy barrier to move between the states. 

\subsection{Design parameters}

The design parameter $\sigma$ is defined as the rest length (state 0) of a muscle. For simplicity, $\sigma$ is also the rest length of the logic core's inputs and the inputs and output of all the gates. The width of all the slabs is $0.5\,\sigma$. For the logic core, $\Delta_{in}=0.5\,\sigma$ and $h=2\,\sigma$ for which $\Delta_{out}=0.56\,\sigma$. The parameter values that decide the input lever dimensions are $(l_{in}, l_{out})=(\sigma,\sigma)$, $(\Delta l_{in}, \Delta l_{out})=(\pm 0.5\,\sigma, 0.5\,\sigma)$ and $l=1.5\,\sigma$, and that decide the output lever dimensions are $(l_{in}, l_{out})=(h,\sigma)$, $(\Delta l_{in}, \Delta l_{out})=(\Delta_{out}, \pm 0.5\,\sigma)$ and $l=2\,\sigma$. The positive or negative sign depends on the gate's signal type, i.e., expansion or contraction, respectively.

\subsection{Simulation method}

The slabs are modeled as rigid bodies. The width and thickness of all slabs are 0.5\,$\sigma$ and 0.01\,$\sigma$, respectively. These are simulated in three dimensions using the simulation toolkit HOOMD-blue (v2.6) \cite{HOOMD-blue,Anderson2008,Glaser2015,Nguyen2011,Phillips2011}. The dynamics of each bar $i$ are simulated using the non-momentum conserving Brownian equation of motion. 
\begin{equation}
    \dot{\pmb{r}_i} = \cfrac{1}{\gamma} \pmb{F}_i + \sqrt{2\cfrac{k_BT}{\gamma}}~\pmb{\eta}(t)
\end{equation}
\begin{equation}
    \dot{\theta_i} = \tau_i/\gamma_r,
\end{equation}
where $\pmb{r}_i$ and $\theta_i$ are the position and orientation of bar $i$ respectively. $\pmb{F}_i$ and $\tau_i$ are the net force and torque on $i$ due to volume exclusion interactions and harmonic bonds. The hinge is modeled by two harmonic bonds at either end with the harmonic coefficient $k_0$ and equilibrium length zero. The universal joint is just a single harmonic bond. For the harmonic bonds of joints with tolerance, the potential is zero up till the tolerance distance between the bond ends. 
\begin{figure}[tbhp]
\centering
\includegraphics[width=1\linewidth]{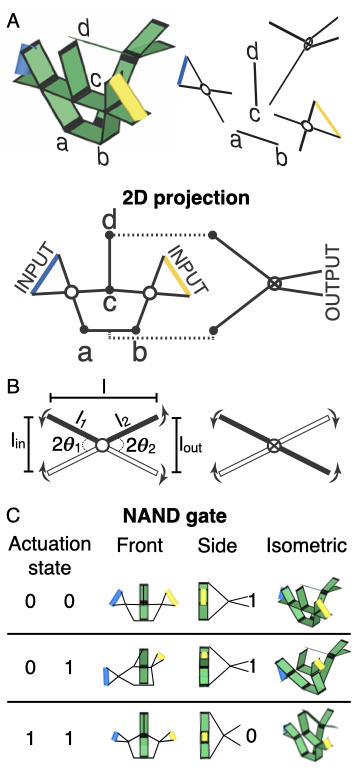}
\caption{NAND gate design. (A) Left is the 3D structure of the NAND gate. Right is the line diagram of this structure illustrating the gate components logic core, two input levers and an output lever. Bottom is the 2D projection of the 3D structure built using the line diagram. The gate structure consists 8 slabs and 8 joints. Seven of these joints are hinges while the one at point $d$ is a universal joint. Dotted lines in the 2D projection represent joints of the output lever to the gate. (B) Two types of scissor lever mechanisms based on their signal inversion behavior. The left one inverts the contraction to expansion and vice-versa whereas the right one retains the actuation direction. For illustration purposes, these are represented by open and crossed hinges respectively. The solid and hollow lines are two rigid bodies connected via the hinge. (C) Truth table of the NAND gate in (A) showing front, side and isometric views of the configurations (simulation snapshots) for different inputs states of the yellow and blue muscles (see Movie S1 for the simulation).}
\label{figure2}
\end{figure}
Since mechanical computation does not require particle collision, volume exclusion is only required to prevent structural overlaps. Hence, we apply isotropic volume exclusion between the centers of the slabs via the Weeks-Chandler-Andersen potential, $U_{WCA}(r)=4\epsilon\left[(2r_0/r)^{12}-(2r_0/r)^6)\right]+\epsilon$, for $r<2r_0$, and $U_{WCA}(r)=0$ otherwise\cite{Chandler1983}, where $r$ is the distance between the centers of the slabs and $r_0=0.1\,\sigma$. Parameter $\epsilon$ determines the strength of the potential and is set as $\epsilon = 10^{-4} \,k_0 \sigma$ for our system. The muscles are modeled as harmonic bonds with strength $k_0$ and are actuated by varying their equilibrium lengths. Thermal energy k$_\text{B}$T = $10^{-5}$\,$k_0 \sigma^2$ unless specified otherwise, and $\pmb{\eta}(t)$ is unit-variance Gaussian white noise. $\gamma_r$ is the rotational drag coefficient and is set equal to the translational drag coefficient, $\gamma$. The rotational noise in a particle orientation arises from the translational diffusion of the particles bonded to it. Time is measured in units of $t_0=\gamma/(10^{-3}k_0)$.

\section{Constructing different gates}

Different gates may use the same logic core because the gate function is decided by the input-output lever designs. Figure 3 illustrates the 2D projection of different gates formed by attaching a logic core to levers that differ in their geometry and mechanism type. For all these gates, except NAND, the muscles contract; the NAND gate uses expanding muscles. The ability of the levers to invert and adapt signals allows them to integrate the NOT gate function within other gates without using additional parts. Whether a lever includes the NOT function depends on how the states are defined. Consider the examples in Figure 3. The OR gate contains a NOT function in all its levers, in contrast to the AND gate, which lacks a NOT function in all levers. The input levers of the OR gate connect the rest state of the muscle (state 0) to the expanded state of the logic core input (state 1). When the muscle contracts to state 1, the crossed hinge mechanism of this lever contracts the logic core input to state 0, thereby integrating the NOT functionality. Similarly, the output lever of the OR gate also includes NOT function. In contrast, the NOR gate's output lever does not include the NOT function. The logic gates are compatible with both expanding and contracting muscles. The NAND gate shown in this figure uses expanding muscles. Hence, for the inputs and outputs of this gate, state 1 is achieved when the rest lengths are expanded by $\sigma/2$. The key difference in this NAND gate from the one in Figure 1 is that the levers' mechanism type is switched. Hence, in this case, the open hinge lever at the logic core's output performs the NOT function. 

\subsection{3D printing of NAND gate}

We 3D printed this NAND gate at the centimeter scale to demonstrate the design, shown in Figure 4. The gate is printed in one piece using a ProJet 3500 model printer that uses M3 Crystal acrylic-based resin as the print material. Components are sized relative to $\sigma$ as discussed in the earlier text, where $\sigma$ = 5\,mm. The thickness of the slabs is 0.1\,mm and the widths are $\sim$\,5\,mm. Rotary hinges are used for the joints with a tolerance of 0.2\,mm between parts. Approximately 0.2\,mm clearance is also left between consecutive knuckles. To build the universal joint at point $d$ (Figure 1A), a separate part with two perpendicular hinges is used (Figure 4). Selective slabs are trimmed to permit rotation at the hinges. Duderstadt Center's Fabrication Studio at the University of Michigan, Ann Arbor, was used for 3D printing.

\begin{figure}[tbhp]
\centering
\includegraphics[width=1\linewidth]{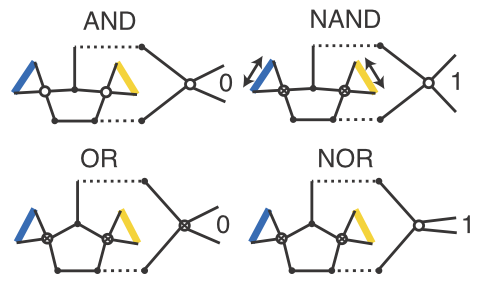}
\caption{2D projections of different logic gates. All the muscles are in their rest state (state 0). All muscles contract by $\sigma/2$ to switch to state 1 except those of the NAND gate, which expand (indicated by the arrow). The bottom dotted line of the gates is simplified as compared to that in Figure 2A. The geometry is drawn to an approximate scale. }
\label{figure3}
\end{figure}

\begin{figure}[tbhp]
\centering
\includegraphics[width=1\linewidth]{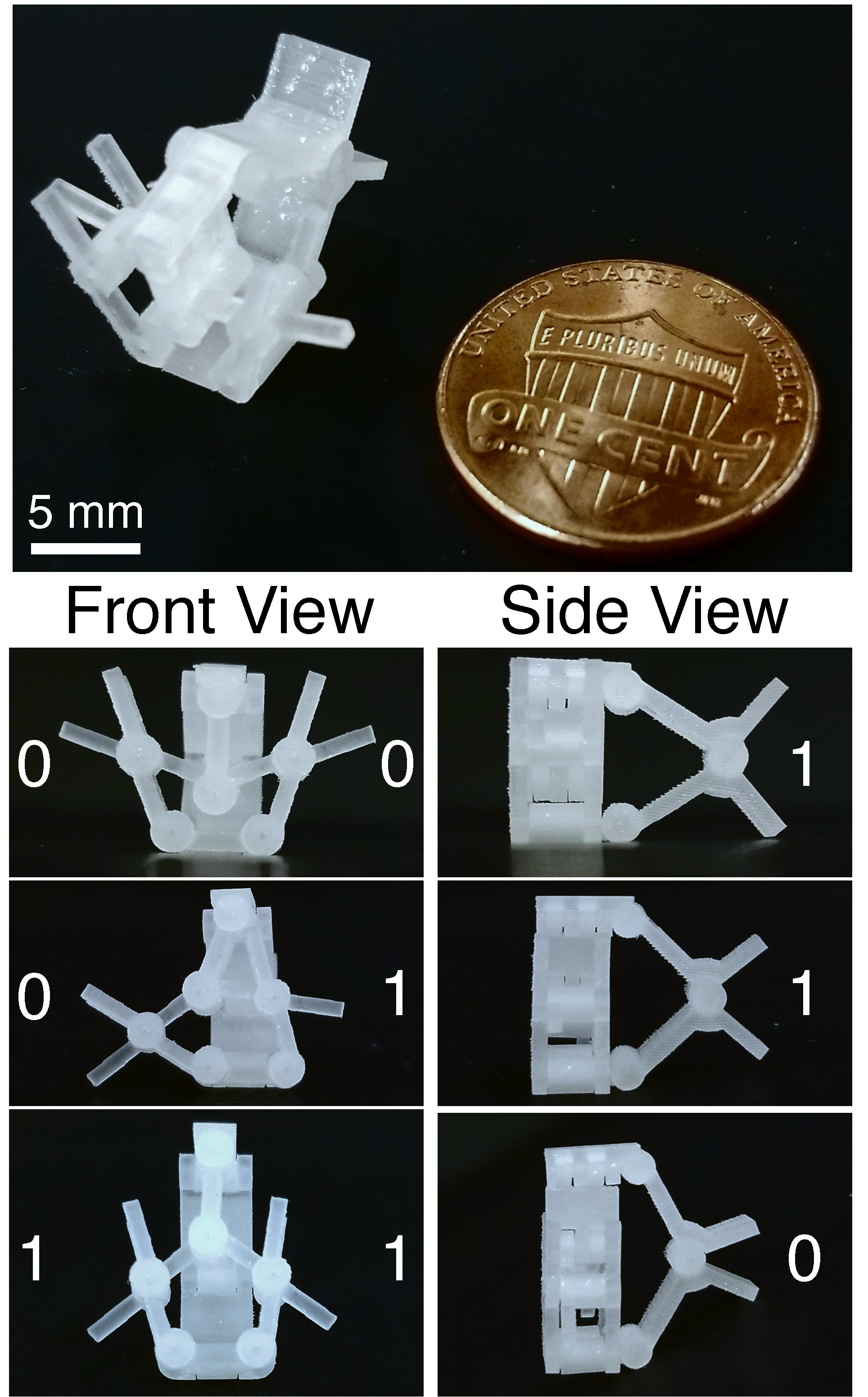}
\caption{3D printed NAND gate. Top is the isometric view of the NAND gate placed near a penny. The gate corresponds to the NAND 2D projection in Figure 3. Bottom is the truth table of the gate. The gate inputs, as shown in the front view, are set manually. }
\label{figure3}
\end{figure}

During the 3D printing, we learned that optimizing for tighter tolerances is difficult when the structure is scaled down. This is because, besides the print resolution, the different relative placement of the hinges with respect to the print plane also governs their tolerance limits. Advancements in printing and lithography techniques have now allowed building micron scale features with precision\cite{Merkel2010,Niesler2014,Ru2014}, which will reduce mechanical tolerances. Alternate ways to improve these tolerances at micron and sub-micron scale is to use flexure joints\cite{Song2019} or single-stranded DNA\cite{Yurke2000,Marras2015,Rogers2011}. See Table 1 for the fabrication techniques corresponding to scales 10\,nm, 1\,$\mu$m and 100\,$\mu$m, along with the related muscle and joint materials, and their strengths.

\begin{figure*}[t]
\centering
\includegraphics[width=17.8cm]{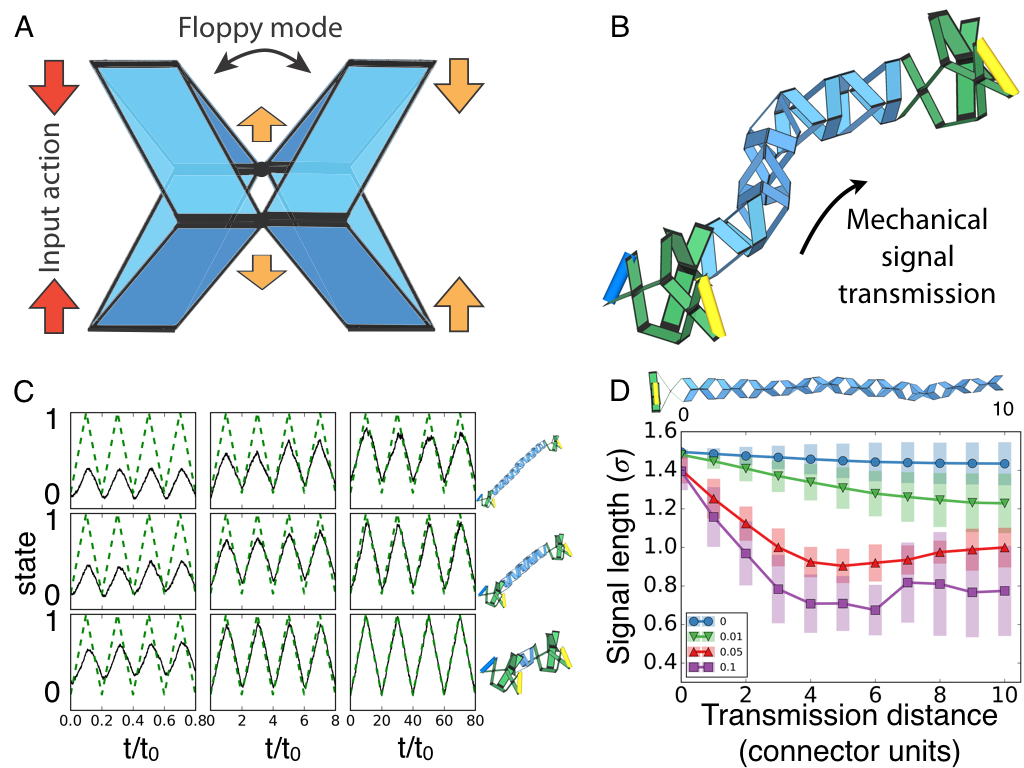}
\caption{Connector design and analysis. (A) A connector unit that transfers the signal and has a floppy mode in the middle. Half unit of the connector, consisting of 4 slabs attached at 4 hinges, inverts the actuation signal and rotates it to its perpendicular axis. The broader black lines are hinge joints. The half units are connected using two universal joints. (B) Simulation snapshot showing transmission of a mechanical signal from the output of the first AND gate to the input of the second. The figure shows signal transmission while allowing the connector chain to bend (see Movie S2 for the simulation). The blue and yellow muscles expand to actuate. (C) Relaxation of the output state of the second gate in (B) is plotted (solid line) for three actuation frequencies (columns) for the connector consisting 1, 5 and 10 units (rows). The dotted line is the actuation cycle of the muscles, which are expanded simultaneously. (D) Signal length, defined as the rest length ($\sigma$) + signal, is plotted against the transmission length, calculated in the number of connector units, for different joint tolerances (indicated by different markers and colors). Tolerance, measured in the unit of $\sigma$, is the distance up to which the jointed slabs are free to separate. The vertical bars at the data points is the variance of the signal length.}
\label{figure5}
\end{figure*}

\section{Connector design}
Since the input-output signals and the rest lengths of different gates are compatible, the output of one gate can be directly connected to the input of another. However, signal transmission may be required for building complex circuits. We propose a design of a connector chain that besides transmitting the signal, can also rotate its actuation axis. A unit of the connector chain is shown in Figure 5A. It consists of two half units with four slabs each. The half unit inverts the actuation signal and rotates the input signal to its perpendicular axis. The full unit transmits the signal but has a floppy mode in this perpendicular axis resulting in a chain with floppy bending modes. Signals can be split by branching connector chains. Figure 5B shows an example of a connector chain connecting the output of an AND gate to the input of another AND gate. The floppy chain results in a flexible circuit. Movie S2 is the simulation of this example showing the simultaneous signal transmission and the bending of the chain. An alternate design for the transmission chain would be to chain together a series of scissor levers. This design is simpler than the one proposed but has the drawback that its chain length varies when actuated. The large chain length variations can impose undesired constraints on the circuits and additional drag forces.

The connectors are simulated using the method described earlier in the text for the logic gates. For this paper, the length of a parallelogram slab of connector is 1.1\,$\sigma$ and is angled at $\sim\,\pi/4$\,radians with respect to the vertical.

\begin{table*}
\centering
\caption{Available technology to enable muscle-inspired transduction at the colloidal scale}
\begin{tabular}{p{1cm} p{1.7cm} p{1cm} p{1.5cm} p{1.5cm} p{1.8cm} p{.1cm} p{1cm} p{1.3cm} p{1.3cm} }
&&\multicolumn{4}{c}{\pmb{Muscle}}&&\multicolumn{3}{c}{\pmb{Hinge}}\\
\cline{3-6}
\cline{8-10}
Muscle length scale & Structure \newline fabrication & Type & Muscle strength & Actuation stimuli & Actuation \newline time scale & & Type & Hinge strength & $^*$Thermal energy over hinge strength (k$_\text{B}$T/k$\sigma^\text{2}$)\\
\midrule
10\,nm & DNA linking\cite{Douglas2009, Rogers2016} & DNA\cite{Yurke2000} & Force to pull apart dsDNA $\sim$15\,pN\cite{Yurke2000} & ssDNA\cite{Yurke2000} & $\sim$13\,s\cite{Yurke2000} & & ssDNA\cite{Yurke2000} & $\sim$15\,pN & 10$^{\text{-2}}$\\

1\,$\mu$m & Self-assembly\cite{Elacqua2017}, Lithography\cite{Merkel2010, Niesler2014}, Non-contact 3D printing\cite{Ru2014} & \cellcolor{blue!10} Stimuli-responsive polymer\cite{Ghosh2017,Hu2019,Hugel2002} & \cellcolor{blue!10}Modulus 100\,kPa - 200\,MPa\cite{Ghosh2017} & \cellcolor{blue!10} Temperature, pH, electricity, light, and chemicals\cite{Stuart2010,Ghosh2017,Hugel2002,Hu2019} & \cellcolor{blue!10}Seconds to hours depending on the scale and material\cite{Hu2019,Stuart2010} & & multiple ssDNA\cite{Marras2015,Rogers2011} & Binding energy per DNA bond 6\,k$_\text{B}$T\cite{Rogers2011} & 10$^{\text{-4}}$ \\

100 $\mu$m & Lithography\cite{Merkel2010, Verotti2015}, 3D printing\cite{Vaezi2013, Niesler2014} & \cellcolor{blue!10} & \cellcolor{blue!10} & \cellcolor{blue!10} & \cellcolor{blue!10} & & Polymer flexure\cite{Song2019, Verotti2015} & -- & --\\
& & Shape-memory alloy\cite{Kim2009} & Modulus of metals >1 GPa & Temperature\cite{Kim2009}, magnetic field\cite{Ghosh2017} & $\sim$10 s\cite{Kim2009} & & Rotary hinge\cite{Pister1992} & Modulus for steel >1 GPa & 10$^{\text{-16}}$\\
\bottomrule
\end{tabular}
\addtabletext{*Appendix provide estimation details.}
\end{table*}

\begin{figure*}[t]
\centering
\includegraphics[width=17.8cm]{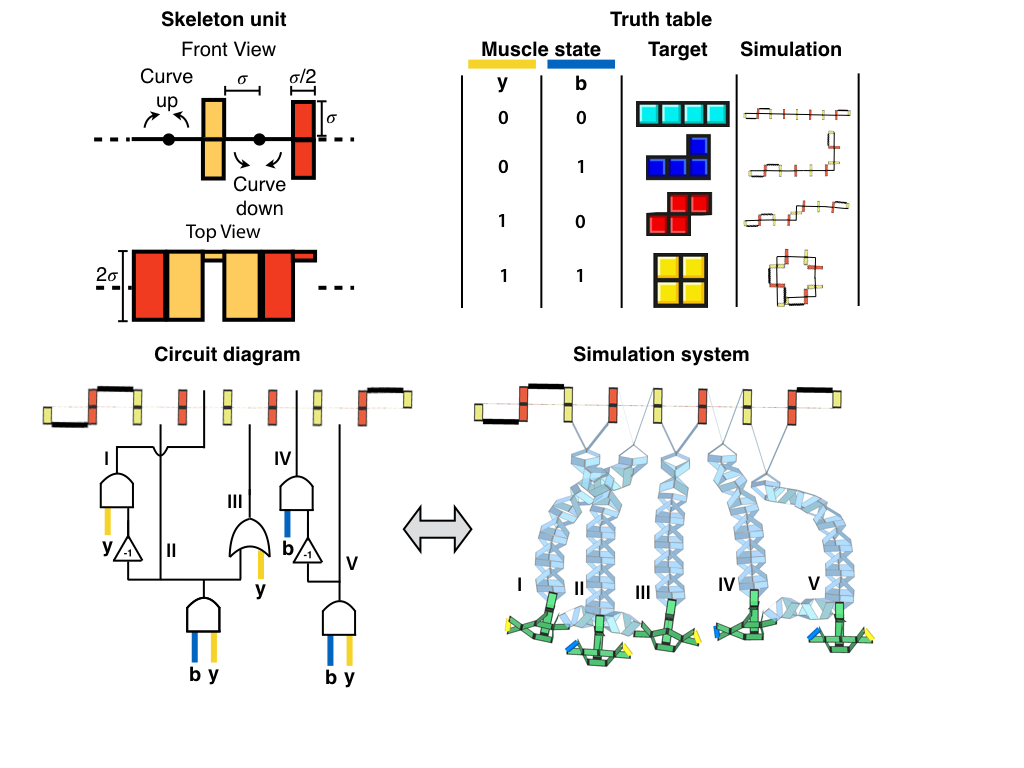}
\caption{Transduction of Tetris shapes. Top-left is the unit of the robot skeleton containing the degrees of freedom to bend upward and downward. The two colors differentiate alternating rigid bodies. Top-right is the truth table with snapshots of 4 Tetris shapes folded in simulations for different states (rest and expanded) of yellow ($y$) and blue ($b$) muscles. Bottom is the logic circuit connected to the skeleton comprised of four units. Bottom-left is the digital representation of the simulation system at bottom-right. Snapshot of the simulation is for the input $(y,b)=(0,1)$. The rest length of the skeleton's degrees of freedom is 2\,$\sigma$ and expands by $\sigma$. Levers connect the circuit to the skeleton via universal joints. See Movies S3-S5 for the simulation.}
\label{figure6}
\end{figure*}

\section{Analysis of a simple circuit}

The two important characteristics of mechanical circuits that govern their efficiency and utility are their response and information loss. We analyze these characteristics for the circuit in Figure 5B. Figure 5C shows the actuation frequency response of this circuit for three different connector chain lengths -- 1, 5 and 10 units -- and for three different frequencies. The response (solid line) is measured as the relaxation of the output state of the second gate as the muscles are actuated and relaxed (dotted line). The root-mean-square deviation of the response from the ideal response (dotted line) for the chain length of 5 units is approximately 17\,\%, 4.5\,\% and 3.6\,\% with decreasing actuation frequency 5\,$t_0^{-1}$, 0.5\,$t_0^{-1}$ and 0.05\,$t_0^{-1}$ respectively. To put the plot in perspective, consider a circuit with micron scale hinges in an aqueous environment. $\gamma$ can be evaluated using the viscosity of water and $k_0$ can be evaluated using the binding energy ($k\sigma^\text{2}$) estimation in Appendix. This estimates $t_0$ to be of the order of a second for micron scale hinges. The time scale of SRPs varies from seconds to hours, which is consistent with the lowest actuation frequency (third column of Figure 5C). The relaxation time will increase with the complexity of the circuit. Using a low drag environment, such as air, and higher joint strength will reduce this time scale. 

Information loss in the mechanical circuits occurs primarily due to joint tolerances and stochastic forces. We quantify this loss in Figure 5D by measuring the signal length (rest length + signal) in a simulation system consisting of an AND gate with a connector at its output. All the joints in this system have a mechanical tolerance, which is the distance up to which the jointed points are free to separate before incurring the energetic penalty due to the bond potential. The reduction in the average signal length along the connector chain is due to mechanical tolerance while the increasing variance in the signal is due to the cumulative effect of thermal stochasticity. Using flexure hinges\cite{Song2019} can lower the mechanical tolerances relative to the rotary hinges. The thermal effect will increase as the system scale is reduced. The information loss due to the thermal stochasticity can be tackled by binary pinning of the states using bistable parts\cite{Song2019,Raney2016}.

\section{Building Tetris robot}

In Figure 6 we demonstrate building a logic circuit and using it to fold a skeleton chain into Tetris shapes. The skeleton is a chain of units, each of which possesses two hinges to bend up and down (see top-left in Figure 6). We choose four Tetris shapes for the skeleton to fold into using two muscles (yellow and blue), which expand when actuated. The truth table (top-right in Figure 6) defines the system states and shows the simulation snapshots of the folded skeleton using the two muscles. This truth table is accomplished by the simulation system (bottom-right in Figure 6) accompanied by its analogous digital circuit (bottom-left in Figure 6). See Movies S3-S5 for the simulation. This simulation system is a proof-of-concept demonstration of processing multiple stimuli and then using the processed information to manipulate separate degrees-of-freedom and achieve an action.

\section{Discussion}

\subsection{Conclusion}

We proposed a design architecture that uses multiple stimuli to trigger the actuation of materials to compute combinatorial logic for miniature scale robotics. The mechanical actuation can be transmitted as a signal between the gates using bendable connectors. The architecture uses only link and joint mechanisms. The gate mechanism continuously transitions between the configurations in a single mode and hence, in principle, does not possess any transition energy barrier. The architecture is stable under stochastic forces relevant to the colloidal scale. The particular novelties we implemented are muscle-like expansion/contraction as the signal, compact gate design lacking sliding or colliding parts, NOT gate integration with the logic gate and a connector design with floppy bending modes. We discussed the mechanism referred to as the logic core, which computes the logic and, with the addition of levers, builds various gates. We 3D printed a cm-scale AND gate as a working prototype. We performed Brownian dynamics simulation to test the architecture at colloidal scales. Through these simulations, we analyzed the signal transmission along the connector length for different mechanical tolerances and the response relaxation for different connector lengths and actuation frequencies. Finally, a complete robot consisting of a mechanical circuit connected to a skeleton chain was simulated to demonstrate, as proof-of-concept, the applicability of our design for robotics. The skeleton of this robot folded into four different Tetris shapes using two different muscles.

\subsection{Outlook and future directions}

Research in mechanical logic is traditionally motivated by either the importance in the fundamental understanding of the relation between logic and physical structure\cite{Prakash2007,Elbaz2010,Woodhouse2017} or due to their potential in building computing systems resilient to extreme conditions of heat, pressure and radiation. We foresee employing mechanical logic in miniature scale robotics for three reasons:\\
1) Mechanical logic is inherently material agnostic and hence the design can be translated to any application context. 
2) Development of materials like SRPs that actuate in response to a variety of stimuli provide transduction routes alternative to electronics at colloidal scales.
3) Modern fabrication techniques can make complex structures at small scales (see Table 1 for the available technologies).

The major criticisms of mechanical computers are that they are slow and prone to wear and tear. However, they also equip conventional robotics with alternative approaches to incorporate "smart" abilities. Using our approach, organic and biomorphic robots can be built that directly harness and transduce the environmental stimuli into action. Material agnosticism allows the usage of biocompatible and biodegradable raw materials. Hence, mechanical logic is advantageous for building robotic systems that require low complexity and speed relative to electronic counterparts, but are constrained in terms of raw material and activation environment. 

The muscle-like actuation we use in our study is precisely defined to expand by or contract to half its length. However, in practice, the SRPs may not actuate precisely. Incorporating bistable elements in the architecture will binary pin the states to increase precision\cite{Song2019,Raney2016}. Bistability can also be used to amplify the applied force of actuation. Bistability along with flexure joints will lower the signal loss due to tolerance limitations and stochastic forces, and speed-up the response. In the future, memory can also be incorporated to enable learning. Memory along with an energy storage mechanism will allow strategic harness and release of the energy from stimuli. Manufacturing of 3D structures with hinges, though possible, is generally difficult. However, circuits can be converted into soft mechanisms, which will be easier to 3D print. Additionally, since the core logic mechanism works in 2D, it may be possible to fabricate an array of these mechanisms on a 2D sheet for computation. The exact structure can be further simplified by optimizing the design for the application in context.

We envision the following applications for our design:
\smallskip

1) {\it Smart medicines:} Nanoscale robotic carriers morph in response to chemical signatures of damaged tissues to release drugs and repair the site.
\smallskip

2) {\it Smart textiles:} Fabric embedded network of tiny mechanical actuators and computers that changes its microstructure tuning optical, electronic, and mechanical properties in response to environmental cues.
\smallskip

3) {\it Space rovers:} Mechanical computers can survive extreme conditions and actuation can directly be powered by local environment, and hence can be potentially used in robots for planetary exploration.
\smallskip

4) {\it Non-invasive surgical robots:} Mechanical circuit mounted on a tentacle on the order of 100 microns can be injected in the human body. An applied magnetic field can navigate and orient the tentacle and lasers can be used to activate specific input muscles. This will in turn actuate the specific configuration of the tentacle to perform surgical operations.
\smallskip

5) {\it Contaminant capture:} Swarms of sub-micron scale robots can detect contaminants with certain chemical signatures and then morph in response to contain and remove them. Such behavior can be used in chemical plants, oil spills, and pipelines.

\section*{Acknowledgments}
This work was supported as part of the Center for Bio-Inspired Energy Science, an Energy Frontier Research Center funded by the U.S. Department of Energy, Office of Science, Basic Energy Sciences under Award \# DE-SC0000989. Our computational resources are supported by Advanced Research Computing at the University of Michigan, Ann Arbor.

\section{Appendix}

\subsection{Estimation of k$_\text{B}$T\,/\,k$\pmb{\sigma}^\text{2}$ in Table 1} 

The thermal energy k$_\text{B}$T$\,=\,$4.1\,pN$\cdot$nm. k$\sigma^\text{2}$ is the binding energy of the joint and is estimated as follows using the hinge strengths given in Table 1.
\smallskip

{\it For 10\,nm:} Applying a force of 15\,pN for a couple of nanometers to break the DNA bond is equivalent to an energy of 15\,pN$\cdot$nm. Assuming $\sim$10 bonds on a 10\,nm wide hinge, k$_\text{B}$T/k$\sigma^\text{2} \approx$ 4.1/15$\cdot$10 $\sim$ 10$^\text{-2}$.
\smallskip

{\it For 1\,$\mu$m:} Assuming $\sim$10$^\text{3}$ DNA bonds on a micron wide hinge gives a net binding energy of $\sim$\,6$\cdot$10$^\text{3}$ k$_\text{B}$T. Thus, k$_\text{B}$T/k$\sigma^\text{2} \approx$ 1/6000 $\sim$ 10$^\text{-4}$.
\smallskip

{\it For 100\,$\mu$m:} Assuming the hinge metal has the strength of $\sim$1GPa. The energy required to pull apart a hinge of area ($\sim$100$\mu$m)$^\text{2}$ by a distance of one $\mu m$ is $\sim$10$^\text{9}$N/m$^\text{2}$$\cdot$10$^\text{-8}$m$^\text{2}$$\cdot$10$^\text{-6}$m = 10$^\text{-5}$N$\cdot$m. Thus, k$_\text{B}$T/k$\sigma^\text{2} \approx$ 4.1$\cdot$10$^\text{-21}$N$\cdot$m/10$^\text{-5}$N$\cdot$m $\sim$ 10$^\text{-16}$.

\section{Supplementary information}
Here is the youtube link to the supplementary videos: \url{https://www.youtube.com/playlist?list=PLft1aMiHLUXRfaP6fA_mIXqNjt8NuddNM}

The video descriptions are as follows:
\smallskip

Movie S1 description: Mechanical NAND gate consisting of only hinges and slabs. Blue and yellow are the muscle-like actuators that contract to go from State 1 to State 0. The mechanical signal is the contraction length.
\smallskip

Movie S2 description: Brownian Dynamics simulation showing transmission of a mechanical signal from the output of the first AND gate to the input of the second. The figure shows signal transmission while allowing the connector chain to bend (see publication for the details). The blue and yellow are muscle-like actuators that expand to go from State 0 to State 1.
\smallskip

Movie S3 description: Proof-of-concept demonstration of folding of a Tetris shape using mechanical logic circuit. Folding was performed via Brownian Dynamics simulation for 1800 t0 (see publication for the details). The blue and yellow are muscle-like actuators that expand to go from State 0 to State 1. The shape is achieved by the expansion of the blue actuator.
\smallskip

Movie S4 description: Proof-of-concept demonstration of folding of a Tetris shape using mechanical logic circuit. Folding was performed via Brownian Dynamics simulation for 1800 t0 (see publication for the details). The blue and yellow are muscle-like actuators that expand to go from State 0 to State 1. The shape is achieved by the expansion of the yellow actuator.
\smallskip

Movie S5 description: Proof-of-concept demonstration of folding of a Tetris shape using mechanical logic circuit. Folding was performed via Brownian Dynamics simulation for 1800 t0 (see publication for the details). The blue and yellow are muscle-like actuators that expand to go from State 0 to State 1. The shape is achieved by the expansion of both blue and yellow actuators.




\clearpage
\bibliography{main}

\end{document}